\documentclass{article}

\usepackage{spconf,amsmath,amssymb,graphicx,booktabs,hyperref,microtype,tikz}
\usetikzlibrary{arrows.meta,positioning,fit,calc}
\hypersetup{hidelinks,hypertexnames=false}
\usepackage{multirow}
\newcommand{\method}{\textsc{BAER}}
\newcommand{\fullname}{Backbone-Adaptive Evidence Routing}

\title{Backbone-Adaptive Evidence Routing for Robust Pairwise LLM Judging}
\name{Zeyan Li$^1$, Jing Peng$^1$, Jianfeng Xu$^1$\sthanks{Corresponding author.}}
\address{$^1$ Shanghai Jiao Tong University}

\begin{document}
\ninept
\maketitle

\begin{abstract}
Pairwise language-model judges can gather evidence through direct comparison, reasoning, or reference-based verification, but no single protocol is best across benchmarks and judge backbones. We introduce \fullname{} (\method{}), which adapts the evidence mechanism while preserving candidate symmetry: swapping the two responses may reverse the preference but cannot change its strength. \method{} separates each expert's signed preference from candidate-invariant reliability and builds three symmetric heads: evidence stacking, reliability-based expert routing, and candidate-blind reference verification. Development data select one head for each benchmark--backbone condition, and that choice is frozen before testing. Across four benchmarks and two 8B judge backbones, \method{} achieves the highest test accuracy among the compared methods in all eight conditions, with full prediction coverage and gains of 0.87--7.32 points over the strongest external baseline. The results show that adapting how evidence is gathered is more reliable than fixing one judging protocol everywhere.
\end{abstract}

\begin{keywords}
LLM-as-a-judge, pairwise evaluation, evidence routing, preference evaluation
\end{keywords}

\section{Introduction}
\label{sec:intro}

Pairwise language-model judges make it practical to compare model outputs at a scale that would otherwise require human raters, and they are now used for model evaluation, translation, and open-ended chat \cite{chiang2023alternative,zheng2023mtbench}. Given an instruction and two candidate responses, the judge states which response is better. Platforms and meta-benchmarks increasingly rely on this two-response interface \cite{chiang2024arena,lambert2025rewardbench,tan2025judgebench}. Choosing how to use the interface is more involved, because a protocol that works well for one task or judge backbone, that is, the language model that performs the judging, may not be the strongest choice for another. We call the combination of a benchmark and a judge backbone a \emph{condition}.

Studies that evaluate the judges themselves have found recurring weaknesses. Judges can be fooled by adversarially constructed responses \cite{zeng2024llmbar}, often prefer whichever response appears in a favored position \cite{wang2024fair,shi2025judging}, and are swayed by surface features such as response length or the tokens used to read out the decision \cite{dubois2024lengthcontrolled,zheng2024pride,li2025calibraeval}. Pairwise judgments can also disagree with an independently constructed basis for evaluation \cite{jeong2025prepair}. A correction designed for one condition may therefore discard useful evidence in another. The problem concerns both how to judge and which evidence to trust in each condition.

\begin{figure}[t]
\centering
\includegraphics[width=\columnwidth]{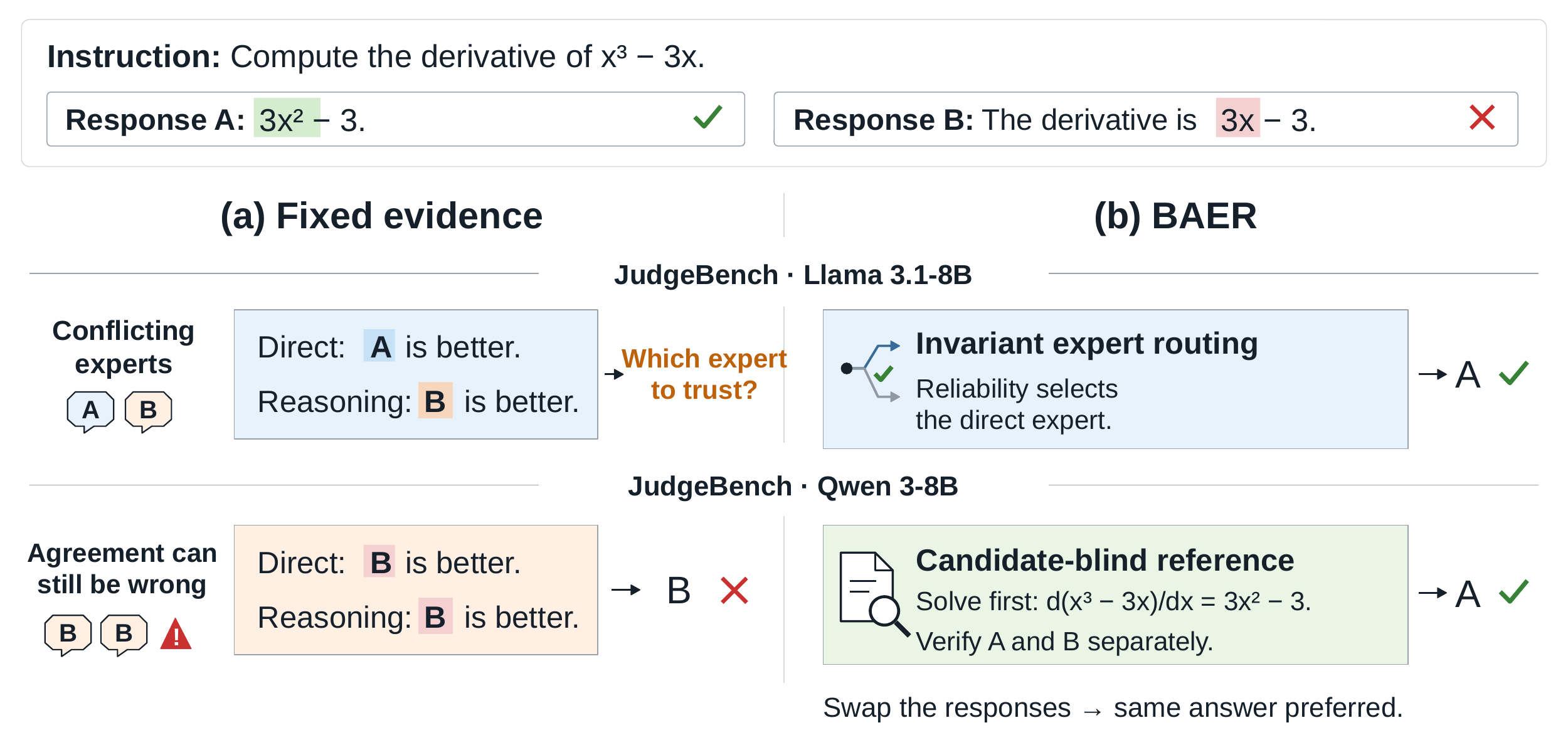}
\caption{Motivation for condition-adaptive evidence. Conflicting experts and shared errors call for different mechanisms, illustrated by BAER's invariant expert routing and candidate-blind reference verification. Head assignments match the deployed JudgeBench conditions; examples are schematic.}
\label{fig:evidence-motivation}
\end{figure}

Figure~\ref{fig:evidence-motivation} illustrates two reasons why the useful evidence mechanism can change. Experts may disagree, or they may agree on an incorrect answer. In the first case, selecting a reliable expert can resolve the disagreement. In the second, combining the same judgments may be insufficient, and a reference constructed without seeing either candidate provides another basis for verification.

Adapting the evidence must not make its selection depend on how the candidates are ordered. A judge whose answer changes when the two responses are swapped is measuring display order rather than quality, so any adaptive mechanism must remain blind to that order. We call each raw judging procedure, such as direct comparison or chain-of-thought, a \emph{protocol}, and its output an \emph{expert signal}. We separate that signal into a signed preference, which records which response the expert favors and how strongly, and a candidate-invariant reliability, which records how trustworthy the expert is on the current pair regardless of order. Swapping the responses reverses the first and leaves the second unchanged. This separation underlies \method{} and its three alternative evidence paths, which we call \emph{heads}. A stack combines the expert signals, a router selects one expert for each pair, and reference verification checks the candidates against a separately constructed solution. Development data determine which head to deploy for each condition. The deployed head remains fixed at test time, while expert selection within the routing head varies from pair to pair.

Existing methods mainly improve a fixed judging protocol. One line changes how a comparison is elicited through explicit reasoning, rubric-style evaluation, or repeated sampling \cite{wei2022cot,liu2023geval,wang2023selfconsistency}. A second keeps the protocol fixed and calibrates a predetermined decision rule against known biases \cite{zheng2024pride,li2025calibraeval}. A third aggregates a fixed panel of judges \cite{verga2024poll}. Selective methods decide when to trust a judgment and when to abstain \cite{badshah2026scope}, with related work providing statistical tools for risk-controlled selection \cite{angelopoulos2025ltt}. None of these addresses the case in which the useful evidence mechanism itself changes across conditions. Aggregation cannot repair an error shared by all pairwise experts, while abstention gives up coverage exactly on uncertain pairs. \method{} instead asks which symmetric evidence head should be deployed in each condition while still producing a prediction for every pair.

Two adaptation scales are deliberately separated. Head selection is \emph{condition-level}: development data choose stacking, routing, or reference verification for a benchmark--backbone pair, and the choice is frozen before test labels are observed. Only the routing head adapts \emph{within} a condition, selecting an expert for each pair from candidate-invariant reliability features. This distinction prevents test-time head shopping and prevents the router from exploiting display order. It also makes failures interpretable: a weak condition can require a different evidence source even when individual experts remain useful on particular pairs.

We make three contributions. First, we formulate evidence adaptation in a way that preserves candidate symmetry. Second, we implement three evidence heads that select heads at the condition level and experts at the sample level. Third, we empirically study when each head helps. Across eight benchmark--backbone conditions, \method{} achieves the highest test accuracy among the compared methods, with margins over the strongest external baseline ranging from +0.87 to +7.32 points. The development experiments further show when stacking, routing, and reference verification provide useful evidence.

\section{BAER}
\label{sec:method}

\begin{figure*}[t]
\centering
\includegraphics[width=\textwidth]{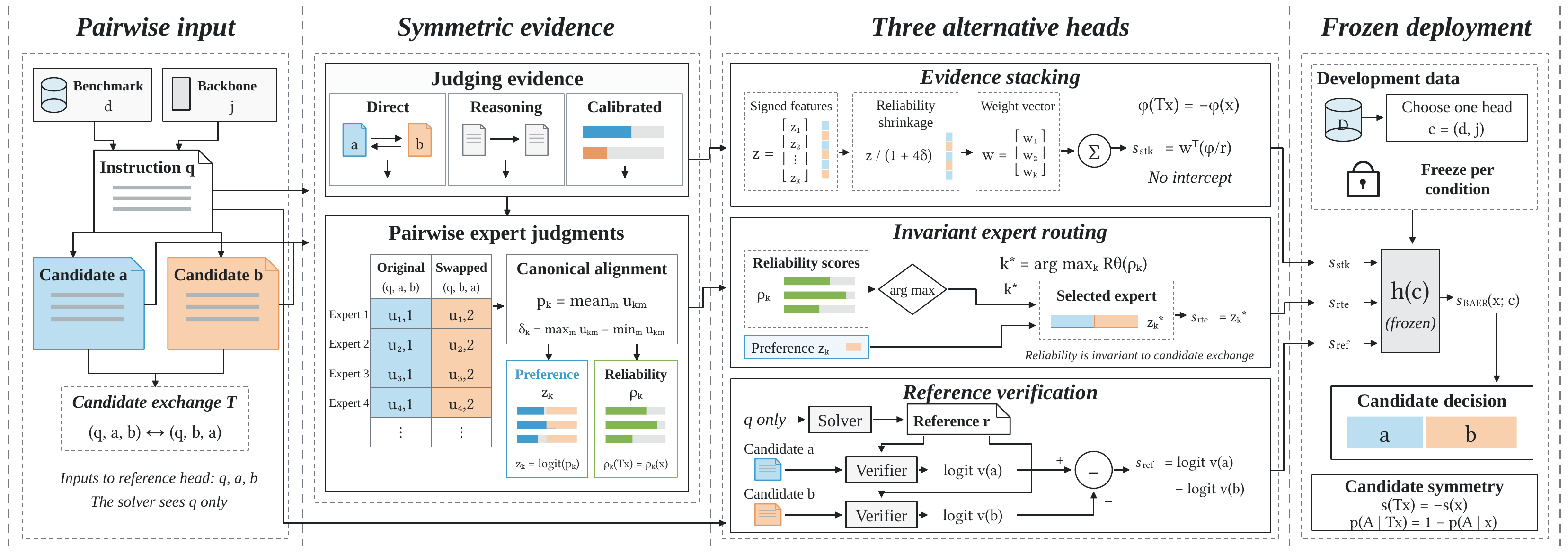}
\caption{Overview of BAER. Development data freeze one evidence head per benchmark--backbone condition. Stacking combines signed protocol features, routing selects one expert per pair using candidate-invariant reliability, and reference verification solves the instruction without seeing either candidate, then checks each independently. All heads return a candidate-symmetric signed score.}
\label{fig:baer-architecture}
\end{figure*}

\method{} adapts how preference evidence is gathered while treating the two candidates symmetrically. As shown in Figure~\ref{fig:baer-architecture}, all heads share one output format, a signed score whose sign picks the winner and whose magnitude measures confidence, and swapping the candidates flips that sign. We first define the score interface and evidence representation, then describe the three heads and the deployment rule.

For instruction $q$ and candidates $a,b$, let $x=(q,a,b)$ and encode the benchmark preference as $y\in\{-1,+1\}$, where $+1$ means that $a$ is preferred. All \method{} heads return a signed score through the same interface,
\begin{equation}
p_h(A\mid x)=\sigma(s_h(x)),\qquad
\widehat y_h=2\mathbb{1}[s_h(x)\geq0]-1 ,
\label{eq:interface}
\end{equation}
where $\mathbb{1}[\cdot]$ is the indicator function. Candidate exchange is $Tx=(q,b,a)$. We require
\begin{equation}
s_h(Tx)=-s_h(x), \qquad p_h(A\mid Tx)=1-p_h(A\mid x).
\label{eq:symmetry}
\end{equation}
Swapping the responses may flip the judgment, but it does not change its strength.

\subsection{Symmetric evidence representation}

The expert bank contains nine judging protocols. These are direct bidirectional judging, self-consistency \cite{wang2023selfconsistency}, chain-of-thought \cite{wei2022cot}, rubric-based judging, score-then-choose judging, PRePair \cite{jeong2025prepair}, PriDe \cite{zheng2024pride}, CalibraEval \cite{li2025calibraeval}, and an internal deliberate A versus B expert with constrained extraction. Each expert may be run under both display orders, with every probability mapped back to the canonical candidate $a$, so a probability always refers to the same response regardless of display position. Let $u_{km}(x)$ be the $m$-th such probability for expert $k$, and let $M_k$ be the number of runs. Its directional confidence and order instability are
\begin{equation}
p_k=\frac{1}{M_k}\sum_{m=1}^{M_k}u_{km},\qquad
\delta_k=\max_m u_{km}-\min_m u_{km}.
\label{eq:evidence-summary}
\end{equation}
Here $p_k$ is the expert's average preference for $a$, and $\delta_k$ is large for an expert whose answer moves across repeated or swapped runs. We form raw and reliability-shrunk log odds,
\begin{equation}
z_k=\operatorname{logit}(p_k), \qquad
\widetilde z_k=\frac{z_k}{1+4\delta_k}.
\label{eq:features}
\end{equation}
The shrinkage downweights experts that order instability identifies as unreliable. Missing evidence is assigned $p_k=.5$, hence zero signed evidence. Under exchange, $p_k\mapsto1-p_k$ and $\delta_k$ is unchanged, so both $z_k$ and $\widetilde z_k$ negate. With $\bar z=K^{-1}\sum_kz_k$, the stack feature map is
\begin{equation}
\begin{aligned}
\phi(x)&=\big[\{z_k,\widetilde z_k\}_{k=1}^{K},
 \{\mathbb{1}[g(x)=c]\bar z\}_{c\in\mathcal G}\big],\\
\phi(Tx)&=-\phi(x).
\end{aligned}
\label{eq:stack-features}
\end{equation}
Here $\mathcal G$ is the set of subset and task-type categories, such as chat, safety, reasoning, and math. The second block interacts the mean evidence with indicators for these categories, letting the stack weight experts differently across task types. These categories are candidate-invariant and are fitted without test labels.

\subsection{Backbone-adaptive evidence heads}

\textbf{Evidence stacking.}
The stack is a logistic model over these features. Each dimension is divided by its training RMS $r_\ell$, computed over the training split, so that experts with naturally larger scores do not dominate. With the sign-augmented set $\mathcal D^{\pm}$, obtained by adding $(-\phi_i,-y_i)$ for every $(\phi_i,y_i)$, \method{} fits
\begin{equation}
\begin{aligned}
w_\lambda&=\arg\min_w\frac{1}{|\mathcal D^{\pm}|}
 \sum_{(v,t)\in\mathcal D^{\pm}}\ell\!\left(tw^\top(v/r)\right)
 +\frac{\lambda}{2}\lVert w\rVert_2^2,\\
s_{\rm stk}(x)&=w_\lambda^\top(\phi(x)/r),
\qquad \ell(u)=\log(1+e^{-u}).
\end{aligned}
\label{eq:stack-objective}
\end{equation}
This augmentation forces the learned model to be odd in $\phi$, and the stack has no intercept, so antisymmetry holds by construction. We select $\lambda\in\{10,1,0.1,0.01,0.001\}$ on calibration data and refit on all non-test rows. The stack feature dimension is 44 for RewardBench, 36 for JudgeBench, 20 for HH-RLHF, and 28 for UltraFeedback, with selected $\lambda$ values $0.001/0.01$, $10/0.01$, $0.01/0.01$, and $0.001/0.01$ for Qwen and Llama respectively.

\textbf{Candidate-invariant expert routing.}
Averaging can erase a strong specialist with votes from experts that know nothing about the pair. Routing replaces the average with a learned selector. For pair $i$ and expert $k$, let $\widehat y_{ik}$ be the binary prediction of expert $k$, and let $t_{ik}=\mathbb{1}[\widehat y_{ik}=y_i]$ state whether that expert is correct. An auxiliary model $R_\theta$ estimates each expert's reliability on the current pair:
\begin{equation}
\begin{aligned}
R_\theta(\rho_{ik})&\approx P(t_{ik}=1\mid\rho_{ik}),\\
k^*(x)&=\arg\max_kR_\theta(\rho_k(x)),\\
s_{\rm rte}(x)&=z_{k^*(x)}(x).
\end{aligned}
\label{eq:router}
\end{equation}
The vector $\rho_{ik}$ contains expert identity, absolute confidence, cross-expert agreement, $\delta_k$, confidence-distribution statistics, symmetric response-length and structure features, and subset identity. All entries are candidate-invariant, so $\rho_k(Tx)=\rho_k(x)$. Exchange preserves the selected expert $k^*$ and negates only the selected score $z_{k^*}$. Logistic regression and tree ensembles are compared by deterministic five-fold cross-validation; both deployed routers select a depth-5 random forest \cite{breiman2001random} with minimum leaf size 8 and refit it on all non-test rows. JudgeBench/Llama also uses an isolated pointwise expert with score
\begin{equation}
s_{\rm pnt}=\operatorname{logit}P(Y\mid q,a)
-\operatorname{logit}P(Y\mid q,b),
\label{eq:pointwise}
\end{equation}
where each probability is normalized over constrained one-token $Y/N$ outputs. Neither verifier sees the other candidate, and this pointwise expert is available only as an additional candidate for the routing head.

\textbf{Candidate-blind reference verification.}
When all experts make the same error, no combination or selection of their judgments can recover. Reference verification adds evidence of a different kind by comparing each candidate with an independently constructed solution. JudgeBench/Qwen 3-8B uses three calls. The judge first sees only $q$ and produces a reference solution $r$, then evaluates each candidate independently under the same instruction and reference, with one-token Y/N constrained decoding. Let $\pi_l(t)=P(l\mid q,r,t)$ for $l\in\{Y,N\}$. Then
\begin{equation}
\begin{aligned}
v(t)&=\frac{\pi_Y(t)}{\pi_Y(t)+\pi_N(t)},\\
s_{\rm ref}(x)&=\operatorname{logit}v(a)-\operatorname{logit}v(b).
\end{aligned}
\label{eq:reference}
\end{equation}
The solver never sees either candidate, and each verifier sees exactly one, so swapping candidates negates the score. The prompt, 4096-token solution budget, constrained decoding, and aggregation are frozen before test.

\subsection{Condition-level selection and deployment}

Recall that a condition $c=(d,j)$ combines a benchmark $d$ and a judge backbone $j$. All pairs in a condition use the same head, so deployment never uses test information per example. Stacking is the initial choice, and development evaluations assess alternatives where it performs poorly. Cross-validation selects the expert router, while a head that relies on reference verification must pass its selection and calibration margins. The resulting condition sets $\mathcal C_{\rm stk}$, $\mathcal C_{\rm rte}$, and $\mathcal C_{\rm ref}$ are disjoint and fixed for final scoring:
\begin{equation}
s_{\method}(x;c)=
\begin{cases}
s_{\rm stk}(x),&c\in\mathcal C_{\rm stk},\\
s_{\rm rte}(x),&c\in\mathcal C_{\rm rte},\\
s_{\rm ref}(x),&c\in\mathcal C_{\rm ref}.
\end{cases}
\label{eq:dispatch}
\end{equation}
Because the condition does not depend on candidate order, $c(Tx)=c(x)$. Since every branch is odd, $s_{\method}(Tx;c)=-s_{\method}(x;c)$. Within a routing condition, $k^*(x)$ in Eq.~\ref{eq:router} can vary across pairs, but candidate exchange preserves both the head and the expert.

\section{Experiments}
\label{sec:experiments}

\subsection{Setup}

We use four preference benchmarks, RewardBench filtered \cite{lambert2025rewardbench} (chat, safety, reasoning), JudgeBench \cite{tan2025judgebench} (knowledge, reasoning, math, coding), HH-RLHF \cite{bai2022hh} (helpfulness and harmlessness), and UltraFeedback \cite{cui2024ultrafeedback} (highest- versus lowest-rated non-tied completions). Their selection/calibration/test sizes are 588/1,234/1,163, 136/238/246, 1,717/3,335/3,500, and 12,598/25,406/25,599, respectively, with original labels. The judge backbones are Qwen 3-8B \cite{qwen3} and Llama 3.1-8B Instruct \cite{llama3}, run locally with greedy decoding except for five-trial self-consistency. Every pair is judged in both response orders and mapped back to the canonical frame, so accuracy reflects content rather than display position. We compare the eight external protocols of Section~\ref{sec:method} on matched test IDs under their native inference procedures. This measures attainable accuracy without controlling for inference cost, and BAER's expert portfolio uses more calls than a single-pass judge. The metric is full-partition pairwise accuracy,
\begin{equation}
\operatorname{Acc}=\frac{1}{N}\sum_{i=1}^{N}\mathbb{1}[\widehat y_i=y_i].
\label{eq:accuracy}
\end{equation}
Missing or unparsed outputs count as errors. We report prediction coverage, defined as the fraction of test pairs for which a method produces a prediction, and use the exact two-sided McNemar test \cite{mcnemar1947test} against each column's strongest external baseline. Development used only development data, first for stacking, then for routing, and finally for reference verification when JudgeBench/Qwen remained weak. Each new head was tuned on the selection and calibration partitions and run once on test inputs with labels withheld.

\subsection{Overall effectiveness}

\begin{table*}[t]
\centering
\small
\setlength{\tabcolsep}{2pt}
\renewcommand{\arraystretch}{0.96}
\caption{Full-test pairwise accuracy (\%) on the benchmark--backbone columns. Bold marks the best result, underline marks the strongest external baseline, and $^{\dagger}$ marks a significant exact two-sided McNemar comparison against the strongest baseline in that column ($p<.05$).}
\label{tab:main}
\begin{tabular*}{\textwidth}{@{\extracolsep{\fill}}lcccccccc@{}}
\toprule
\multirow{2}{*}{Method} & \multicolumn{2}{c}{RewardBench} & \multicolumn{2}{c}{JudgeBench} & \multicolumn{2}{c}{HH-RLHF} & \multicolumn{2}{c}{UltraFeedback} \\
\cmidrule(lr){2-3} \cmidrule(lr){4-5} \cmidrule(lr){6-7} \cmidrule(lr){8-9}
 & Qwen 3-8B & Llama 3.1-8B & Qwen 3-8B & Llama 3.1-8B & Qwen 3-8B & Llama 3.1-8B & Qwen 3-8B & Llama 3.1-8B \\
\midrule
Direct bidirectional & 81.9 & 75.2 & 58.5 & 49.6 & 60.1 & 56.8 & 91.2 & 62.3 \\
Self-consistency & 81.9 & 75.5 & 58.5 & 50.0 & 60.1 & 57.1 & 91.2 & 62.3 \\
Chain-of-thought & 82.3 & 62.9 & \underline{66.7} & 37.4 & 59.6 & 53.3 & 89.2 & 56.5 \\
Rubric-based & 40.7 & 66.7 & 31.3 & 24.8 & 51.9 & 50.5 & 45.0 & 49.4 \\
Score-then-choose & 42.3 & 68.9 & 36.6 & 34.1 & 49.1 & 52.8 & 45.0 & 52.0 \\
PRePair & 69.3 & 62.2 & 53.3 & 44.7 & 54.8 & 54.1 & 70.8 & 63.1 \\
PriDe & 78.8 & 73.3 & 57.7 & \underline{50.4} & 57.1 & 57.7 & 89.3 & 63.7 \\
CalibraEval & \underline{88.7} & \underline{93.4} & 56.9 & 48.4 & \underline{87.3} & \underline{92.5} & \underline{97.6} & \underline{87.7} \\
\midrule
\textbf{BAER (ours)} & \textbf{91.6}$^{\dagger}$ & \textbf{99.3}$^{\dagger}$ & \textbf{70.7} & \textbf{57.7} & \textbf{91.1}$^{\dagger}$ & \textbf{99.7}$^{\dagger}$ & \textbf{98.5}$^{\dagger}$ & \textbf{94.9}$^{\dagger}$ \\
\bottomrule
\end{tabular*}
\end{table*}

\method{} achieves the highest accuracy in every Table~\ref{tab:main} column, with margins from +0.87 to +7.32 points (mean +4.90). The strongest external method changes across columns. CalibraEval leads six columns, while chain-of-thought and PriDe lead the two JudgeBench columns. This is direct evidence that no fixed protocol dominates, which is the variation \method{} exploits through condition-specific heads. Six of the eight gains are significant at the 5\% level under an exact two-sided McNemar test \cite{mcnemar1947test}. The two non-significant comparisons are the two JudgeBench columns. \method{} also produces a prediction for every test pair. In contrast, chain-of-thought has coverage between 74.8\% and 100\% across columns.

\subsection{Contributions of the evidence heads}

Stacking provides the broad foundation. Routing and reference verification improve the three conditions where it is weaker. Column counts track the progression. The constrained A/B expert alone leads 1/8 of the columns, the stack-based system 5/8, the routing-augmented version 7/8, and full \method{} 8/8. Selecting the routing head raises JudgeBench/Llama from 47.6\% to 57.7\% and UltraFeedback/Qwen from 97.0\% to 98.5\%. Replacing the previous router with reference verification raises JudgeBench/Qwen from 60.2\% to 70.7\%.

The transfer behavior explains this division of roles. Six stacks remain within 1.5 points from calibration to test, while the two JudgeBench stacks drop 9.5 and 9.1 points. The harder JudgeBench pairs therefore need evidence of a different kind. Routing is itself condition-sensitive. The former JudgeBench/Qwen router gained 4.8 out-of-fold points but reached only 60.2\% on test, and simple confidence switching and subset lookup also fail. An any-expert oracle, which uses the correct label to select the best expert per pair and so measures the available headroom, reaches 95.4\%/96.2\%. The correct judgments are present inside the expert bank. Identifying them from reliability features, however, remains difficult.

\begin{figure}[!t]
\centering
\includegraphics[width=\columnwidth]{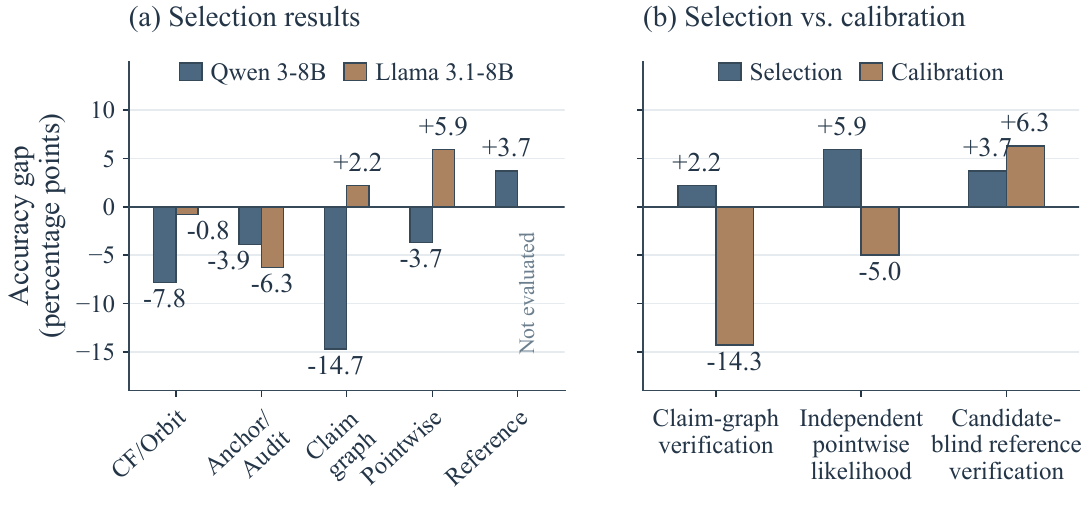}
\caption{JudgeBench semantic-head development. Accuracy gaps are relative to the strongest external baseline on matched IDs; zero denotes parity. (a) Selection results for the best variant per family ($n=128$; other heads use $n=136$ and are named in (b)). CF = counterfactual reconciliation, Orbit = complete orbit, Anchor = blind anchor, Audit = independent pointwise audit. (b) Selection and calibration ($n=238$) for the three advancing heads. Only candidate-blind reference verification retains a positive gap; the Llama reference head was not evaluated.}
\label{fig:judgebench-development}
\end{figure}

\subsection{From selection to calibration}

A development head advances only if its gains survive two gates, a small selection split and a larger calibration split. The JudgeBench development experiments in Figure~\ref{fig:judgebench-development} show why both are needed. The 128-example screening runs do not pass the selection gate. Claim-graph verification improves by +2.2 points for Llama on selection, then reverses to -14.3 on calibration. Independent pointwise likelihood improves by +5.9 points on selection, then reverses to -5.0 on calibration. Candidate-blind reference verification is the Qwen head that clears both gates, with gains of +3.7/+6.3 points. Advancement is decided by accuracy margin, not by significance. The corresponding McNemar $p$-values at these partition sizes are .551/.096. The deployed solution budget remains 4096 tokens; 82/116 selection/calibration solutions reach this limit. Two further 128-example screening studies test broader replacements. CalibraEval's full fourth orbit, a variant that expands the calibration space, wins only for Qwen on JudgeBench (+1.6), losing 0.8--38.3 points elsewhere. 

\section{Conclusion}
\label{sec:conclusion}

We present \method{}, a judging framework that adapts the evidence mechanism to each benchmark and judge backbone while preserving candidate symmetry. It separates each expert's signed preference from candidate-invariant reliability and builds three heads on that separation: stacking, routing, and candidate-blind reference verification. Across four benchmarks and two 8B judge backbones, \method{} achieves the highest accuracy in all conditions, with margins of +0.87 to +7.32 points over the strongest external baseline and a prediction for every test pair. The experiments show why adaptation matters: stacks transfer well on most conditions, routing helps when specialist judgments can be identified from invariant reliability, and reference verification supplies new evidence when the pairwise expert bank shares an error. The gains come with additional inference cost, and condition-level deployment assumes that the benchmark family and judge backbone are known in advance. Future work should extend head selection to unseen conditions and reduce cost through distillation, expert pruning, cached references, or conditional early exits.

\clearpage
\bibliographystyle{IEEEbib}
\bibliography{refs}

@inproceedings{zheng2023mtbench,
  title     = {Judging {LLM}-as-a-Judge with {MT-Bench} and Chatbot Arena},
  author    = {Zheng, Lianmin and others},
  booktitle = {Advances in Neural Information Processing Systems},
  volume    = {36},
  pages     = {46595--46623},
  year      = {2023},
  doi       = {10.52202/075280-2020},
  url       = {https://proceedings.neurips.cc/paper_files/paper/2023/hash/91f18a1287b398d378ef22505bf41832-Abstract-Datasets_and_Benchmarks.html}
}

@inproceedings{zeng2024llmbar,
  title     = {Evaluating Large Language Models at Evaluating Instruction Following},
  author    = {Zeng, Zhiyuan and Yu, Jiatong and Gao, Tianyu and Meng, Yu and Goyal, Tanya and Chen, Danqi},
  booktitle = {International Conference on Learning Representations},
  year      = {2024},
  url       = {https://openreview.net/forum?id=tr0KidwPLc}
}

@inproceedings{shi2025judging,
  title     = {Judging the Judges: A Systematic Study of Position Bias in {LLM}-as-a-Judge},
  author    = {Shi, Lin and Ma, Chiyu and Liang, Wenhua and Diao, Xingjian and Ma, Weicheng and Vosoughi, Soroush},
  booktitle = {Proceedings of the 14th International Joint Conference on Natural Language Processing and the 4th Conference of the Asia-Pacific Chapter of the Association for Computational Linguistics},
  pages     = {292--314},
  year      = {2025},
  doi       = {10.18653/v1/2025.ijcnlp-long.18},
  url       = {https://aclanthology.org/2025.ijcnlp-long.18/}
}

@inproceedings{tan2025judgebench,
  title     = {{JudgeBench}: A Benchmark for Evaluating {LLM}-Based Judges},
  author    = {Tan, Sijun and others},
  booktitle = {International Conference on Learning Representations},
  year      = {2025},
  url       = {https://openreview.net/forum?id=G0dksFayVq}
}

@article{badshah2026scope,
  title   = {{SCOPE}: Selective Conformal Optimized Pairwise {LLM} Judging},
  author  = {Badshah, Sher and Emami, Ali and Sajjad, Hassan},
  journal = {arXiv preprint arXiv:2602.13110},
  year    = {2026},
  url     = {https://arxiv.org/abs/2602.13110}
}

@inproceedings{lambert2025rewardbench,
  title     = {{RewardBench}: Evaluating Reward Models for Language Modeling},
  author    = {Lambert, Nathan and others},
  booktitle = {Findings of the Association for Computational Linguistics: NAACL 2025},
  pages     = {1755--1797},
  year      = {2025},
  doi       = {10.18653/v1/2025.findings-naacl.96},
  url       = {https://aclanthology.org/2025.findings-naacl.96/}
}

@inproceedings{liu2023geval,
  title     = {{G-Eval}: {NLG} Evaluation Using {GPT-4} with Better Human Alignment},
  author    = {Liu, Yang and Iter, Dan and Xu, Yichong and Wang, Shuohang and Xu, Ruochen and Zhu, Chenguang},
  booktitle = {Proceedings of the 2023 Conference on Empirical Methods in Natural Language Processing},
  pages     = {2511--2522},
  year      = {2023},
  doi       = {10.18653/v1/2023.emnlp-main.153},
  url       = {https://aclanthology.org/2023.emnlp-main.153/}
}

@inproceedings{wang2023selfconsistency,
  title     = {Self-Consistency Improves Chain of Thought Reasoning in Language Models},
  author    = {Wang, Xuezhi and others},
  booktitle = {International Conference on Learning Representations},
  year      = {2023},
  url       = {https://openreview.net/forum?id=1PL1NIMMrw}
}

@article{qwen3,
  title   = {{Qwen3} Technical Report},
  author  = {Yang, An and Li, Anfeng and Yang, Baosong and others},
  journal = {arXiv preprint arXiv:2505.09388},
  year    = {2025},
  url     = {https://arxiv.org/abs/2505.09388}
}

@article{llama3,
  title   = {The {Llama 3} Herd of Models},
  author  = {Grattafiori, Aaron and Dubey, Abhimanyu and Jauhri, Abhinav and others},
  journal = {arXiv preprint arXiv:2407.21783},
  year    = {2024},
  url     = {https://arxiv.org/abs/2407.21783}
}

@article{angelopoulos2025ltt,
  title   = {Learn then Test: Calibrating Predictive Algorithms to Achieve Risk Control},
  author  = {Angelopoulos, Anastasios N. and Bates, Stephen and Cand{\`e}s, Emmanuel J. and Jordan, Michael I. and Lei, Lihua},
  journal = {The Annals of Applied Statistics},
  volume  = {19},
  number  = {2},
  pages   = {1641--1662},
  year    = {2025},
  doi     = {10.1214/24-AOAS1998},
  url     = {https://doi.org/10.1214/24-AOAS1998}
}

@inproceedings{zheng2024pride,
  title     = {Large Language Models Are Not Robust Multiple Choice Selectors},
  author    = {Zheng, Chujie and Zhou, Hao and Meng, Fandong and Zhou, Jie and Huang, Minlie},
  booktitle = {International Conference on Learning Representations},
  year      = {2024},
  url       = {https://openreview.net/forum?id=shr9PXz7T0}
}

@inproceedings{li2025calibraeval,
  title     = {{CalibraEval}: Calibrating Prediction Distribution to Mitigate Selection Bias in {LLM}s-as-Judges},
  author    = {Li, Haitao and others},
  booktitle = {Proceedings of the 63rd Annual Meeting of the Association for Computational Linguistics (Volume 1: Long Papers)},
  pages     = {16537--16552},
  year      = {2025},
  doi       = {10.18653/v1/2025.acl-long.808},
  url       = {https://aclanthology.org/2025.acl-long.808/}
}

@inproceedings{jeong2025prepair,
  title     = {The Comparative Trap: Pairwise Comparisons Amplifies Biased Preferences of {LLM} Evaluators},
  author    = {Jeong, Hawon and Park, ChaeHun and Hong, Jimin and Lee, Hojoon and Choo, Jaegul},
  booktitle = {Proceedings of the 8th BlackboxNLP Workshop: Analyzing and Interpreting Neural Networks for NLP},
  pages     = {79--108},
  year      = {2025},
  doi       = {10.18653/v1/2025.blackboxnlp-1.5},
  url       = {https://aclanthology.org/2025.blackboxnlp-1.5/}
}

@article{bai2022hh,
  title   = {Training a Helpful and Harmless Assistant with Reinforcement Learning from Human Feedback},
  author  = {Bai, Yuntao and Jones, Andy and Ndousse, Kamal and others},
  journal = {arXiv preprint arXiv:2204.05862},
  year    = {2022},
  url     = {https://arxiv.org/abs/2204.05862}
}

@inproceedings{cui2024ultrafeedback,
  title     = {{UltraFeedback}: Boosting Language Models with Scaled {AI} Feedback},
  author    = {Cui, Ganqu and others},
  booktitle = {Proceedings of the 41st International Conference on Machine Learning},
  series    = {Proceedings of Machine Learning Research},
  volume    = {235},
  pages     = {9722--9744},
  year      = {2024},
  url       = {https://proceedings.mlr.press/v235/cui24f.html}
}

@inproceedings{wei2022cot,
  title     = {Chain-of-Thought Prompting Elicits Reasoning in Large Language Models},
  author    = {Wei, Jason and others},
  booktitle = {Advances in Neural Information Processing Systems},
  volume    = {35},
  pages     = {24824--24837},
  year      = {2022},
  doi       = {10.52202/068431-1800},
  url       = {https://proceedings.neurips.cc/paper/2022/hash/9d5609613524ecf4f15af0f7b31abca4-Abstract-Conference.html}
}

@inproceedings{dubois2024lengthcontrolled,
  title     = {Length-Controlled {AlpacaEval}: A Simple Way to Debias Automatic Evaluators},
  author    = {Dubois, Yann and Liang, Percy and Hashimoto, Tatsunori B.},
  booktitle = {First Conference on Language Modeling},
  year      = {2024},
  url       = {https://openreview.net/forum?id=CybBmzWBX0}
}

@inproceedings{wang2024fair,
  title     = {Large Language Models Are Not Fair Evaluators},
  author    = {Wang, Peiyi and others},
  booktitle = {Proceedings of the 62nd Annual Meeting of the Association for Computational Linguistics (Volume 1: Long Papers)},
  pages     = {9440--9450},
  year      = {2024},
  doi       = {10.18653/v1/2024.acl-long.511},
  url       = {https://aclanthology.org/2024.acl-long.511/}
}

@inproceedings{chiang2023alternative,
  title     = {Can Large Language Models Be an Alternative to Human Evaluations?},
  author    = {Chiang, Cheng-Han and Lee, Hung-yi},
  booktitle = {Proceedings of the 61st Annual Meeting of the Association for Computational Linguistics (Volume 1: Long Papers)},
  pages     = {15607--15631},
  year      = {2023},
  doi       = {10.18653/v1/2023.acl-long.870},
  url       = {https://aclanthology.org/2023.acl-long.870/}
}

@inproceedings{chiang2024arena,
  title     = {Chatbot Arena: An Open Platform for Evaluating {LLM}s by Human Preference},
  author    = {Chiang, Wei-Lin and others},
  booktitle = {Proceedings of the 41st International Conference on Machine Learning},
  series    = {Proceedings of Machine Learning Research},
  volume    = {235},
  pages     = {8359--8388},
  year      = {2024},
  url       = {https://proceedings.mlr.press/v235/chiang24b.html}
}

@article{verga2024poll,
  title   = {Replacing Judges with Juries: Evaluating {LLM} Generations with a Panel of Diverse Models},
  author  = {Verga, Pat and others},
  journal = {arXiv preprint arXiv:2404.18796},
  year    = {2024},
  url     = {https://arxiv.org/abs/2404.18796}
}

@article{breiman2001random,
  title   = {Random Forests},
  author  = {Breiman, Leo},
  journal = {Machine Learning},
  volume  = {45},
  number  = {1},
  pages   = {5--32},
  year    = {2001},
  doi     = {10.1023/A:1010933404324},
  url     = {https://doi.org/10.1023/A:1010933404324}
}

@article{mcnemar1947test,
  title   = {Note on the Sampling Error of the Difference Between Correlated Proportions or Percentages},
  author  = {McNemar, Quinn},
  journal = {Psychometrika},
  volume  = {12},
  number  = {2},
  pages   = {153--157},
  year    = {1947},
  doi     = {10.1007/BF02295996},
  url     = {https://doi.org/10.1007/BF02295996}
}

\end{document}